# A Clinically Interpretable Deep CNN Framework for Early Chronic Kidney Disease Prediction Using Grad-CAM-Based Explainable AI

Anas Bin Ayub
*Department of Computer Science and Engineering*
*Manarat International University*
Dhaka, Bangladesh
abaanas.ayub2002@gmail.com

Nilima Sultana Niha
*Department of Computer Science and Engineering*
*Manarat International University*
Dhaka, Bangladesh
nilimasultananiha@gmail.com

Md. Zahurul Haque
*Department of Computer Science and Engineering*
*Manarat International University*
Dhaka, Bangladesh
jahurulhaque@manarat.ac.bd

*Abstract*– **Chronic Kidney Disease (CKD) constitutes a major global medical burden, marked by the gradual deterioration of renal function, which results in the impaired clearance of metabolic waste and disturbances in systemic fluid homeostasis. Owing to its substantial contribution to worldwide morbidity and mortality, the development of reliable and efficient diagnostic approaches is critically important to facilitate early detection and prompt clinical management. This study presents a deep convolutional neural network (CNN) for early CKD detection from CT kidney images, complemented by class balancing using Synthetic Minority Over-sampling Technique (SMOTE) and interpretability via Gradient-weighted Class Activation Mapping (Grad-CAM). The model was trained and evaluated on the *CT KIDNEY DATASET*, which contains 12,446 CT images, including 3,709 cyst, 5,077 normal, 1,377 stone, and 2,283 tumor cases. The proposed deep CNN achieved a remarkable classification performance, attaining 100% accuracy in the early detection of chronic kidney disease (CKD). This significant advancement demonstrates strong potential for addressing critical clinical diagnostic challenges and enhancing early medical intervention strategies.**

## I. INTRODUCTION

Chronic Kidney Disease (CKD) has emerged as a silent but formidable epidemic in the twenty-first century, transcending geographical boundaries to become a leading cause of morbidity and mortality worldwide. While the global prevalence is estimated at approximately 10% to 14%, affecting over 850 million individuals, the burden is disproportionately distributed, weighing heavily on Low- and Middle-Income Countries (LMICs) [1]. In these regions, the intersection of resource-constrained healthcare infrastructures, rising metabolic risk factors, and unique environmental nephrotoxins creates a public health crisis of significant magnitude. The progressive nature of CKD, characterized by the gradual loss of renal function leading to End-Stage Renal Disease (ESRD), necessitates early and accurate diagnostic intervention to prevent the catastrophic physiological and economic consequences of kidney failure.

Kidney disease is a major yet under-recognized public health crisis and causes more deaths each year than breast cancer or prostate cancer. In the United States, approximately 35.5 million adults—more than 1 in 7 individuals (14%)—are estimated to have kidney disease [2]. Alarmingly, about 90% of affected adults remain unaware of their condition, as symptoms often do not appear until the disease has progressed to advanced stages, even among individuals with severe kidney disease, nearly 1 in 3 (40%) are unaware of their diagnosis [3]. Furthermore, about 33% of U.S. adults are at risk of developing kidney disease. Currently, kidney disease ranks as the 8th leading cause of death in the country [4].

Most CKD patients in Bangladesh are diagnosed at advanced stages, significantly increasing the demand for dialysis and kidney transplantation. Each year, approximately 35,000–40,000 new cases of kidney failure are reported, while more than 20,000 patients with chronic kidney failure die annually due to their inability to afford dialysis or kidney transplantation [5].

The kidneys are vital organs of the human body, and their proper functioning is essential for maintaining overall health. The kidneys remove wastes and extra water to make urine and filter about a half cup of blood in every minute in the human body [6]. Kidneys also remove acid that is produced by the cells of our body and maintain a healthy balance of water, salts, and minerals such as sodium, calcium, phosphorus, and potassium in our blood [7]. Chronic kidney disease involves a gradual loss of kidney function. Kidneys filter wastes and excess fluids from your blood, which are then removed in your urine. Advanced chronic kidney disease can cause dangerous levels of fluid, electrolytes and wastes to build up in our body [8].

Chronic kidney disease can develop for many reasons, but diabetes and hypertension are the two most common causes. About 25% of kidney failure cases are linked to diabetes, while hypertension accounts for roughly 33% of cases. People with long-term conditions such as hypertension, diabetes, or chronically high blood pressure are at risk of developing serious, life-threatening CKD. Smoking can cause many cardiovascular diseases that can often lead to CKD, quitting smoking, stopping drinking alcohol, eating healthy, exercising regularly, and being careful about taking painkillers can prevent CKD from happening [9]. Chronic kidney disease is a worldwide health crisis. For example, in the year 2005, there were approximately 58 million deaths worldwide, with 35 million attributed to chronic disease, according to the World Health Organization [10].

In the age of Artificial Intelligence (AI) and Machine Learning (ML), researchers are developing innovative approaches to detect CKD by combining traditional clinical diagnosis with advanced algorithms and predictive models [11]. Various models and studies have been conducted and found that most of the researchers used the Conventional Neural Network (CNN) model, which did not perform well on multi-class image classification [12].

After reviewing extensive research in the field, we propose a model for the early detection of Chronic Kidney Disease (CKD) with improved accuracy. In this study, we developed and trained a custom Convolutional Neural Network (CNN) to classify CKD for early detection, achieving 100% accuracy on both the training and testing datasets. Furthermore, to enhance interpretability, we incorporated an Explainable AI (XAI) technique using Grad-CAM to provide visual explanations of the model's predictions.

## II. LITERATURE REVIEW

This section provides a review of recent research studies in the field of image processing. It examines various methodologies and techniques that have been proposed for utilizing image-processing approaches in the diagnosis of kidney diseases. The discussion highlights key strategies, innovations, and contributions presented across the selected works.

Md Nazmul Islam et al. [13], to enable early and accurate diagnosis of kidney diseases, the development of an AI-based diagnostic system is essential. This study focuses on three major categories of renal diseases: kidney stones, cysts, and tumors. A total of 12,446 CT whole abdomen and urogram images were utilized to develop and evaluate the AI-based diagnostic system. Analysis of the dataset revealed a consistent mean color distribution across all classes of images. To identify the most effective approach, six machine learning models were implemented, including EANet, CCT, Swin Transformer, ResNet, VGG16, and Inception v3. The evaluation results indicate that both VGG16 and CCT models achieve satisfactory accuracy, while the Swin Transformer model outperforms the others, attaining an accuracy of 99.30%.

Kim et al. [14] employed an artificial neural network (ANN) to classify chronic kidney disease. Parameters from each region of interest (ROI) were extracted using the gray-level co-occurrence matrix (GLCM) technique, which is widely used in ultrasound image processing. The ANN model consisted of 58 input parameters, ten hidden layers, and three output layers. The model achieved a classification accuracy of 95.4% after 38 training epochs. To further improve performance, they plan to implement a transfer learning approach and expand the dataset for training.

Muneer Majid et al. [15] explored transfer learning techniques for classifying kidney diseases using CT images. To improve the training efficiency, they applied various pre-processing steps and image scaling methods. Within this setup, they proposed two enhanced transfer learning models, ResNet-101 and DenseNet-121, for kidney tumor prediction. Which achieved an impressive accuracy of 94.09%. However, the study had some limitations regarding comparative analysis, model evaluation, and generalizability. In particular, the comparison of detection methods—including Random Forest, Support Vector Machine, Gradient Boosting, Light Gradient Boosting, and deep learning models ResNet-101 and DenseNet-121—was somewhat limited and did not fully consider other relevant models.

Bhandari *et al.* [16] introduced a lightweight convolutional neural network (CNN) for kidney disease detection. They applied the LIME (Local Interpretable Model-agnostic Explanations) method, using the mean and standard deviation of the training data to generate relevant features and modifications. LIME provided a visual explanation of the model's decisions, highlighting the most critical regions of the images for predicting specific classes. The proposed CNN model demonstrated excellent performance, achieving an accuracy of 98.68%. However, the study was conducted on a limited set of CT scans, suggesting that the results could be further improved through data augmentation. Moreover, combining deep learning models with additional explainable AI (XAI) techniques could enhance the interpretability of the results.

Guozhen Chen *et al.* [17] introduced an efficient approach for early detection of Chronic Kidney Disease (CKD) using various deep learning techniques, specifically through an Adaptive Hybridized Deep Convolutional Neural Network (AHDCNN). The study highlighted that to achieve high classification accuracy, a Conventional Neural Network (CNN) model was developed, effectively handling feature dimensions in the dataset. Furthermore, the Internet of Medical Things (IoMT) platform demonstrated that machine learning techniques can provide solutions not only for kidney disease but also for other medical conditions. The proposed system achieved an accuracy of 97.3% in CKD detection.

Fuzhe Ma *et al.* [18] reported that the prevalence of chronic kidney disease (CKD) is steadily increasing, and machine learning techniques have become crucial for its diagnosis in recent years. They proposed a model based on a Heterogeneous Modified Artificial Neural Network (HMANN) for detecting, segmenting, and diagnosing CKD within the Internet of Medical Things (IoMT) framework. The model integrates a Support Vector Machine (SVM) and a Multilayer Perceptron (MLP) trained using the Backpropagation (BP) method. During preprocessing, ultrasound images are utilized for segmentation. The proposed approach not only reduces processing time but also achieves an accuracy of 97.5%.

Swapnita Srivastava *et al.* [19] developed a computational model for the diagnosis of Chronic Kidney Disease (CKD) using publicly available renal disease datasets. The proposed approach employs a performance-tuning nested framework that optimizes hyperparameters and identifies suitable weights for combining ensembles through a Ranking Weighted Ensemble method. The model achieved an accuracy of 98.75%, demonstrating its potential for use in automated systems for early detection of kidney disease.

Nicholas Heller *et al.* [20] highlighted that many studies have explored the relationship between the geometric and anatomical characteristics of kidney tumors and oncology outcomes. Creating high-quality 3D segmentations is a time-consuming and labor-intensive process, both for the tumors and the kidneys in which they are located. In recent years, deep learning techniques have shown promising results in automated 3D segmentation, but they require a large amount of training data. To promote advancements in automatic segmentation, the International Conference on Medical Image Computing and Computer-Assisted Intervention (MICCAI) organized the Kidney and Kidney Tumor Segmentation Challenge (KiTS19) in 2019. In this challenge, 90 cases were evaluated based on the average Sørensen–Dice coefficient for kidney and tumor segmentation. The winning team established a benchmark for 3D semantic segmentation, achieving an accuracy of 97.4% for the kidney and 85.1% for the tumor.

Singh et al. [21] introduced a framework for the early detection of chronic kidney disease using a deep neural network. Their study applied the Recursive Feature Elimination method to identify the most important features contributing to accurate predictions. Various features were then used as inputs to several machine learning classifiers. The proposed deep neural network outperformed models such as SVM, KNN, Logistic Regression, Random Forest, and Naive Bayes. However, a major limitation of their work was the small dataset used for validation. The authors emphasized that, in the future, larger and higher-quality CKD datasets will be needed to more effectively assess disease severity.

Chin-Chi Kuo *et al.* [22] introduced a deep learning–based framework for the automated estimation of glomerular filtration rate (eGFR) and the classification of Chronic Kidney Disease (CKD). Their study utilized 4,505 kidney ultrasound images annotated with corresponding eGFR values. To enhance predictive performance, the authors employed a neural network architecture leveraging transfer learning with a ResNet model pretrained on the ImageNet dataset. Kidney length annotations were incorporated to exclude peripheral regions, ensuring that the model focused on clinically relevant structures. Additionally, multiple data augmentation strategies were applied to increase dataset variability and improve feature extraction. Bootstrap aggregation was further implemented to boost model robustness and minimize overfitting. The proposed AI-GFR model achieved an accuracy of 85.6%, demonstrating strong potential for supporting CKD detection in clinical settings.

Md. Arifuzzaman et al. [23] evaluate several state-of-the-art transfer learning architectures—including EfficientNetV2, InceptionNetV2, MobileNetV2, and ViT—for early CKD detection using a publicly available dataset named CT KIDNEY DATASET: Normal-Cyst-Tumor and Stone, featuring 12,446 unique data. Their findings demonstrate exceptional individual model performance, with ViT achieving 91.5% accuracy and MobileNetV2 surpassing the 90% threshold. To further capitalize on the strengths of each model, the authors propose an ensemble-based diagnostic system that achieves a notable 96% accuracy, outperforming all standalone methods.

Kanwal *et al.* [24] developed an automated framework for kidney abnormality classification, incorporating two advanced deep-learning architectures: EfficientNet-B0 and ResNet-18. Both models demonstrated strong predictive capability in identifying renal pathologies. Among them, the ResNet-18 model achieved the highest performance, attaining an accuracy of 98.1% during evaluation. The authors noted that the next phase of their work will involve collecting real-time datasets to further train and validate their proposed models.

Sudharson *et al.* [25] employed an ensemble-based deep learning framework using transfer learning for kidney ultrasound image classification. Multiple pre-trained DNN architectures—including ResNet-101, ShuffleNet, and MobileNetV2—were used to extract high-level features from three different datasets, after which a support vector machine classifier was applied. Final decisions were generated through a majority-voting fusion strategy across the ensemble models. Their method achieved a maximum classification accuracy of 96.54% on high-quality images and 95.58% on noisy images, demonstrating strong robustness to image variations.

## III. METHODOLOGY

### A. Dataset
1) *Data Collection:* The dataset, titled "CT KIDNEY DATASET: Normal-Cyst-Tumor and Stone" [26], was carefully curated from a range of clinical sources, including multiple hospitals across Dhaka, Bangladesh. It comprises CT scans from patients who had been previously diagnosed with various kidney-related conditions, ensuring broad coverage of pathological scenarios. In total, the dataset contains 12,446 samples, consisting of 3,709 cyst cases, 5,077 normal cases, 1,377 stone cases, and 2,283 tumor cases. To enhance its diagnostic diversity, the dataset includes both contrast-enhanced and non-contrast CT studies, as well as coronal and axial views, providing a robust and representative collection of kidney imaging data.

### B. Data Pre-processing
1) *Image Augmentation:* A key component of this research is the strategic use of data augmentation techniques, including rescaling images to 28×28 based on the model's input requirements. These augmentation methods help diversify the dataset and improve the model's robustness to the natural variations present in kidney images.
2) *Color Space Conversion:* As OpenCV loads images in the BGR (Blue–Green–Red) format, all input CT images were converted to the RGB (Red–Green–Blue) color space using 'cv2.cvtColor()'. This step ensures consistency with deep learning frameworks that are typically trained on RGB images.
3) *Label Encoding:* A label encoding step was applied to transform the categorical class names into numerical identifiers required for supervised learning. The mapping used was: Cyst = 0, Normal = 1, Stone = 2, and Tumor = 3. This numeric representation enables efficient computation and model interpretation.

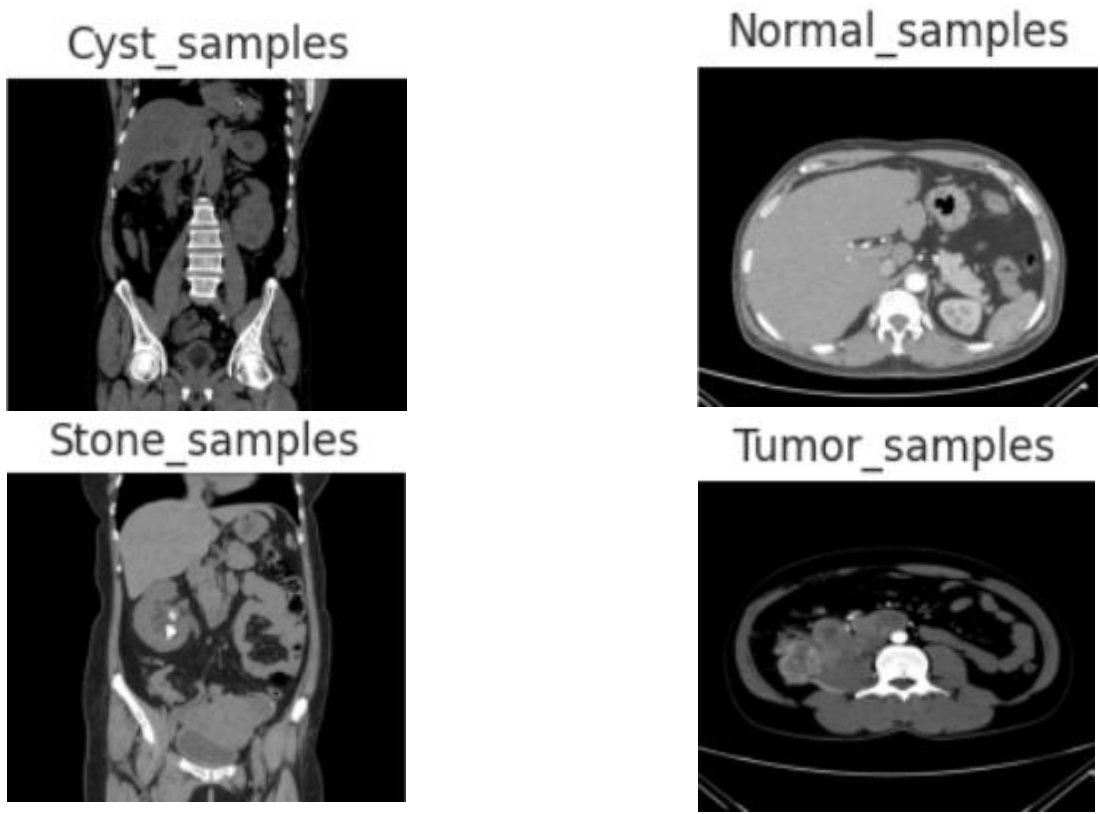

Fig. 1. Random Images from Dataset

4) *Data Balancing Using SMOTE:* To address the class imbalance present in the dataset, the Synthetic Minority Oversampling Technique (SMOTE) was applied. Since SMOTE operates on a two-dimensional feature space, all images were first reshaped from their original 4D tensor structure into a 2D feature matrix. SMOTE then generated synthetic samples for the minority classes by interpolating between existing feature vectors in the corresponding class neighborhoods. This process effectively increased the representation of underrepresented classes while preserving the overall data distribution, resulting in a balanced dataset suitable for robust model training.

5) *Restoring Image Dimensions After SMOTE Oversampling:* After applying SMOTE, the oversampled data existed in a flattened two-dimensional feature matrix, as required for the oversampling process. To prepare the data for input into the deep learning model, the feature vectors were reconstructed into a 4D tensor of size (number of samples, height, width, channels)

6) *Train-Test Split:* To evaluate the generalization performance of the models, an 80–20 train–test split was employed. This partitioning strategy provides a sufficiently large training set to facilitate effective model learning, while preserving an independent test set for objective performance assessment. The 80–20 split offers a practical balance between promoting model convergence and mitigating the risk of overfitting.

7) *Train–Validation Split:* A 90–10 train–validation split was performed using the 'train_test_split' method from scikit-learn., where 90% of the balanced data was used for training and 10% for validation. A fixed random state (42) was set to maintain reproducibility in the dataset partitioning.

C. Convolution Neural Network (CNN) Models

1) *Model Building:* A convolutional neural network (CNN) architecture was implemented using the TensorFlow Keras framework for multi-class kidney abnormality classification. The network comprises three convolutional layers with 28, 64, and 128 filters, respectively, each utilizing a 3×3 kernel with ReLU activation. Max-pooling layers with a 2×2 window follow each convolutional layer to progressively reduce spatial resolution while preserving discriminative features. The resulting feature maps are flattened and fed into a fully connected softmax layer with four units, corresponding to the four diagnostic categories. This architecture offers an efficient balance between representational capacity and computational complexity for CT-image-based kidney analysis.

2) *Model Training:* The convolutional neural network model was trained using the Adam optimizer and the sparse categorical cross-entropy loss function, which is suitable for multi-class classification tasks with integer-encoded labels.

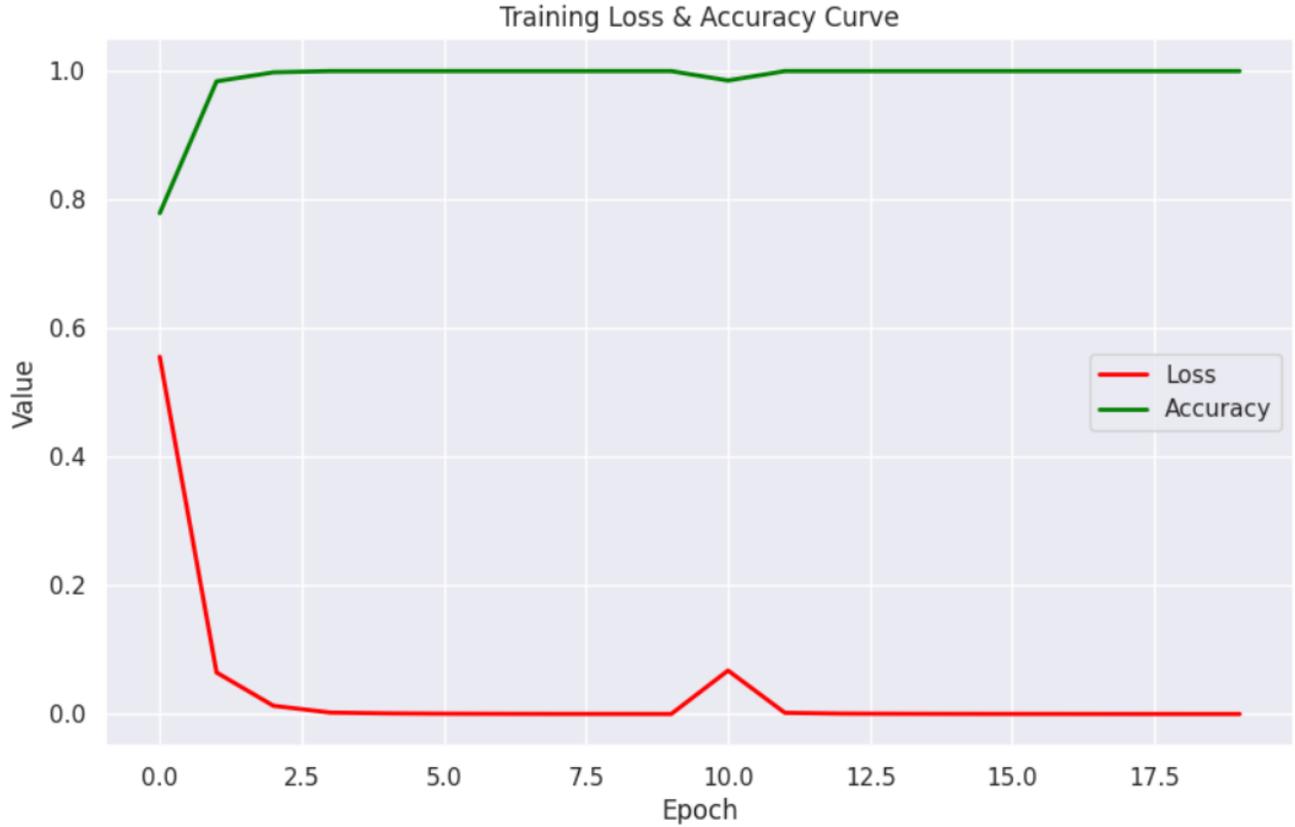

Fig. 2. Training Loss and Accuracy Curve

The model was trained for 20 epochs, where the training set was used to update network weights, and the validation set was employed to monitor generalization performance.

D. *Explainability Using Grad-CAM*

To enhance the interpretability of the proposed deep learning model for kidney disease detection, Gradient-weighted Class Activation Mapping (Grad-CAM) was employed as an Explainable AI (XAI) technique. Grad-CAM generates class-specific heatmaps that highlight the spatial regions of an input image most influential to the model's prediction. By utilizing the gradients flowing into the final convolutional layer to weight the corresponding feature maps, Grad-CAM produces a visual representation of the model's decision-making process. This facilitates better understanding and debugging of the model's behavior and improves transparency and trustworthiness, particularly in medical imaging applications.

IV. PERFORMANCE ANALYSIS

A. *Performance of CNN Models*

The proposed study employed a custom convolutional neural network (CNN) architecture for the identification and classification of kidney diseases using a CT-based kidney dataset. The model was trained using input images resized to 28×28 pixels. Hyperparameters were optimized by setting the batch size to 42 and training the network for 20 epochs.

Owing to the multi-class nature of the classification task, a softmax activation function was incorporated in the output layer. The network was compiled using the Adam optimizer together with the categorical cross-entropy loss function, ensuring effective gradient-based optimization during training.

Five performance metrics - accuracy (ACC), precision (PPR), recall or sensitivity (SEN), F1-score, and the area under the ROC curve (AUC) - were employed across all class datasets to assess the effectiveness of the proposed kidney disease classification model. The class-wise performance metrics for all models are summarized in Table I.

TABLE I. PERFORMANCE ANALYSIS OF CNN MODELS

| Model (Class) | Precision | Recall | F1-Score | AUC |
|---|---|---|---|---|
| Tumor | 1.00 | 1.00 | 1.00 | 1.00 |
| Cyst | 1.00 | 1.00 | 1.00 | 1.00 |
| Normal | 1.00 | 1.00 | 1.00 | 1.00 |
| Stone | 1.00 | 1.00 | 1.00 | 1.00 |

The confusion matrix for CNN models shown in Figure 3, shows that the classifier achieves near-perfect discrimination across all four kidney classes, with predictions concentrated along the diagonal and minimal off-diagonal errors. This indicates consistently high class-wise performance and very low misclassification rates.

The ROC curves shown in Figure 4 for all four kidney classes - tumor, cyst, normal, and stone - exhibit an ideal shape, rising sharply toward the upper-left corner. Each class attains an AUC score of 1.00, indicating perfect separability with zero false-positive and false-negative trade-offs. This demonstrates that the proposed classifier achieves optimal discrimination performance across all categories.

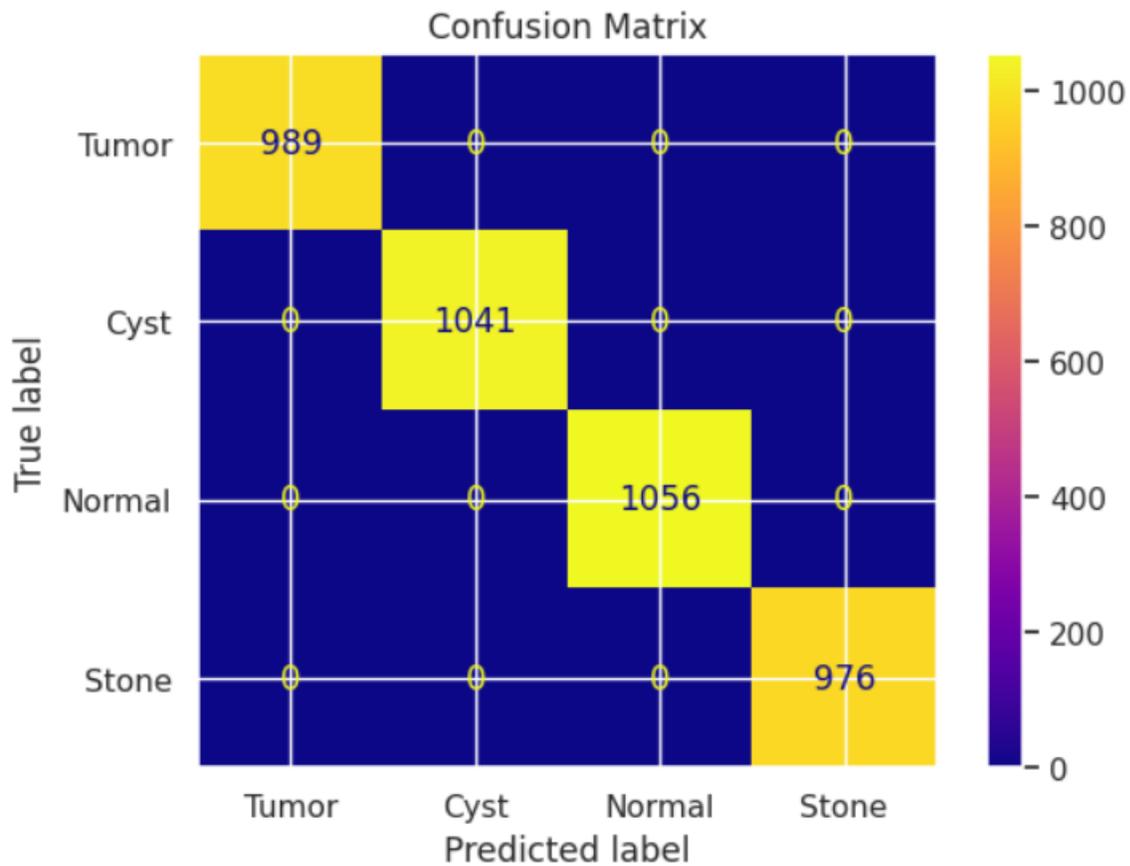

Fig. 3. Confusion Matrix for CNN Models

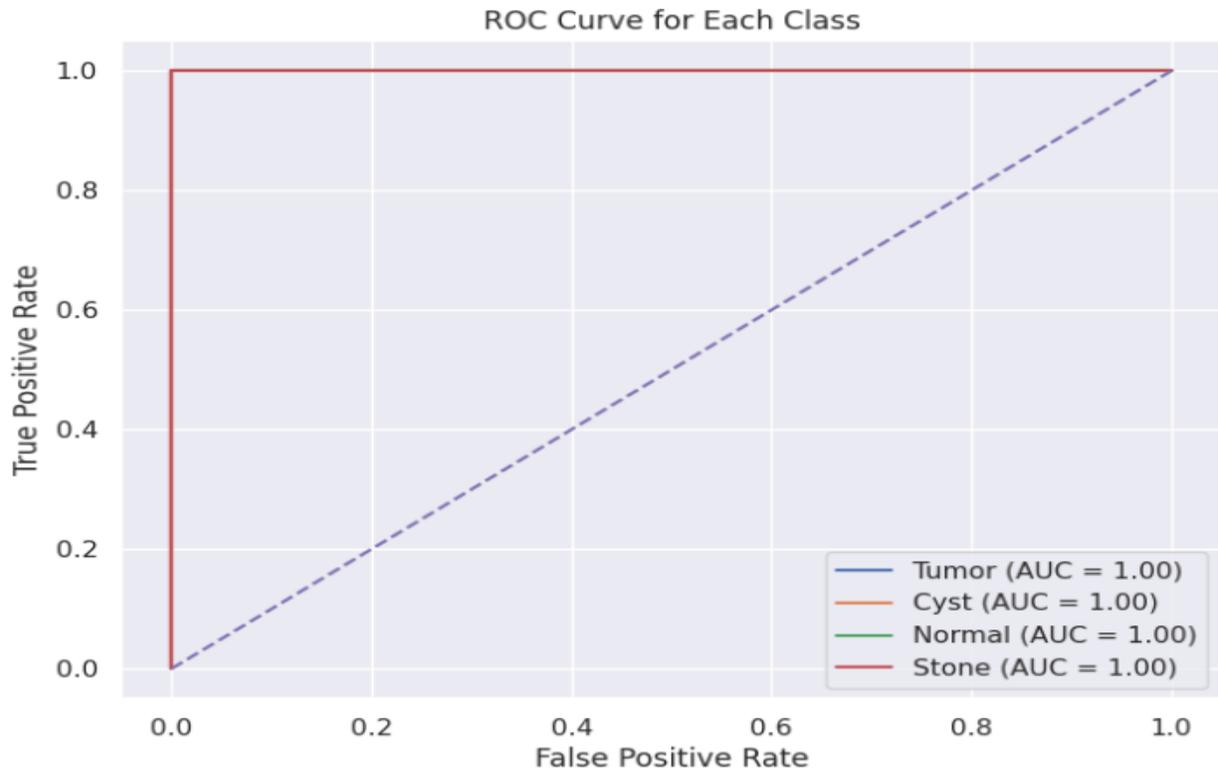

Fig. 4. ROC Curve for Each Class

Precision measures the proportion of correctly identified positive samples (True Positive) among all predicted positives, indicating how reliably the model avoids false positives. Recall quantifies the proportion of actual positive samples correctly detected, reflecting the model's ability to minimize false negatives. These metrics are critical for CNN-based medical classifiers, where both misdiagnosis (low precision) and missed diagnoses (low recall) have serious clinical implications. The Precision–Recall curves shown in figure 5 for all kidney classes - tumor, cyst, normal, and stone - exhibit near-ideal characteristics, maintaining a precision of 1.00 across almost the entire recall range. Each class achieves an Average Precision (AP) score of 1.00, indicating perfect retrieval performance with no trade-off between precision and recall. The flat, upper-bound curves reflect the model's exceptional ability to correctly identify positive samples while avoiding false positives, further confirming the classifier's highly reliable discrimination capability across all categories.

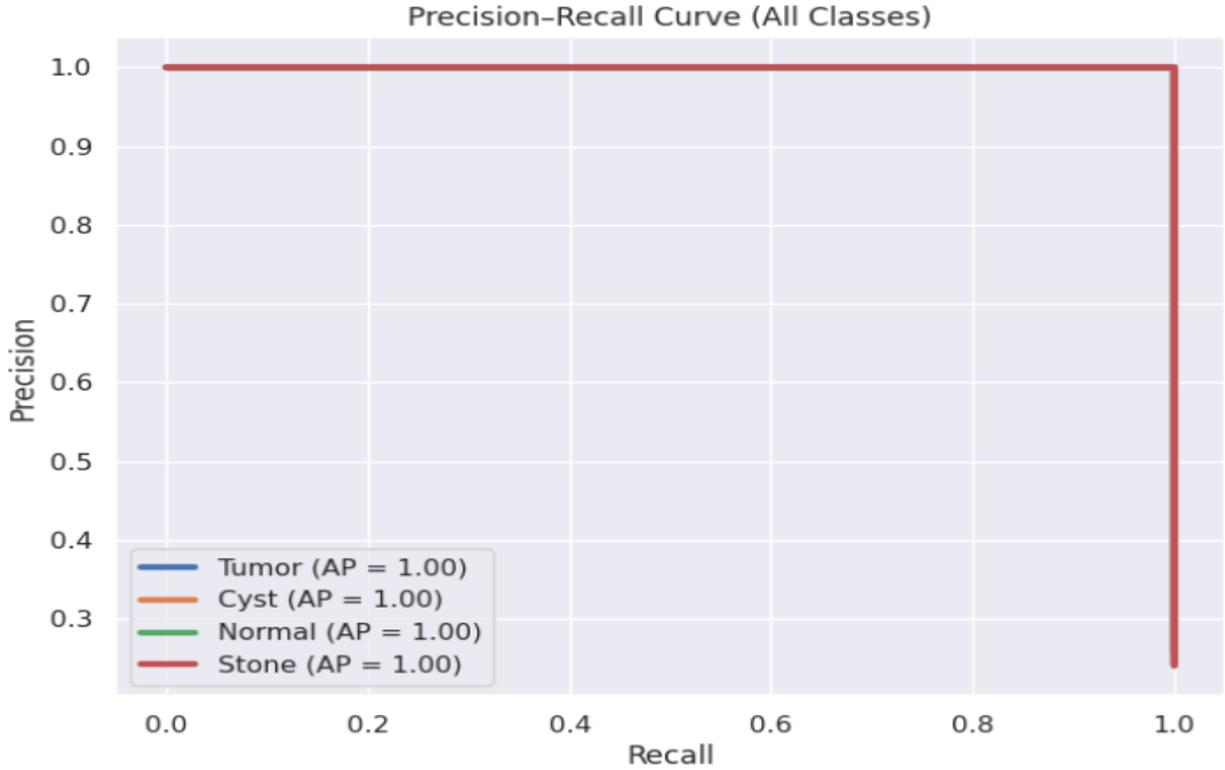

Fig. 5. Precision–Recall Curve for Each Class

B. *Comparison with Existing Model*

A comparative performance assessment between the highest-performing models reported in recent literature and the proposed approach is presented in Table II, where evaluation accuracy serves as the primary benchmark metric. The results emphasize the necessity of systematically evaluating and comparing diverse models and methodological approaches to advance the field and enhance the reliability of image identification and related computational tasks. The CNN models achieved a perfect classification accuracy of 100% and an AUC score of 1.00, underscoring their exceptional discriminative capability. These results highlight the model's strong potential to support clinical decision-making in the prognosis and assessment of kidney diseases.

TABLE II. COMPARISON WITH EXISTING WORKS

| References | Dataset Size | Models Used | Accuracy |
|---|---|---|---|
| Md Nazmul Islam et al. [13] | 12,446 | EANet, CCT, Swin Transformer, ResNet, VGG16, Inception v3 | 99.30% (Swin Transformer) |
| Muneer Majid et al. [15] | 12,446 | ResNet-101, DenseNet-121 | 94.09% (both) |
| Chin-Chi Kuo et al. [22] | 4,505 | ResNet, AI-GFR model | 85.6% |
| Md. Arifuzzaman et al. [23] | 12,446 | Ensemble Method | 96% |
| Fatemeh Zabihollahy et al. [27] | 315 | CNN | 83.75% |
| Devrim Akgun et al. [28] | 460 | MobileNet ResNet50 | 86.42% 82.06% |
| Kalkan et al. [29] | 5000 | ResNet152V2 | 89.58% |

|  |  | MobileNetV2 | 88.80% |
| --- | --- | --- | --- |
| Lee et al. [30] | 1596 | Inception V3 | 74.3% |
|  |  | MobileNet | 72.37% |
| Our Work | 12,446 | Deep-CNN | 100% |

V. CONCLUSION

This study presented a fine-tuned convolutional neural network (CNN) model for the early detection of Chronic Kidney Disease (CKD) using CT kidney images, complemented by SMOTE-based class balancing and Grad-CAM visual explainability. The proposed model achieved a classification accuracy of 100% across all four classes - cyst, normal, stone, and tumor - demonstrating robust discriminative capability and exceptional performance across precision, recall, F1-score, AUC, and PR curves. The incorporation of Grad-CAM further enhanced the interpretability of the model by highlighting anatomically relevant regions that influenced classification decisions, thereby increasing its clinical transparency and trustworthiness.

Compared with state-of-the-art models reported in the literature, the proposed CNN outperformed several deep-learning architectures in terms of classification accuracy and diagnostic reliability. These results indicate that a properly tuned lightweight CNN architecture, coupled with effective balancing and preprocessing strategies, can yield outstanding performance for CT-based kidney abnormality detection. Overall, this work demonstrates strong potential for supporting early diagnosis and improving clinical workflow efficiency in renal disease screening.

Future work will focus on extending the system toward real-world clinical integration. First, we plan to deploy the model within large-scale laboratory decision-support environments to evaluate operational performance. The system will also be expanded to assist in automated e-prescription generation. Incorporating additional clinical metadata - such as eGFR, age, blood pressure, and laboratory profiles - may further improve diagnostic accuracy. Broader validation using multi-center and cross-hospital datasets will be conducted to assess generalizability. Finally, we will focus on developing a mobile or Internet of Medical Things (IoMT)-based platform for remote and point-of-care CKD screening.